\documentclass[sigplan]{acmart}
\setcopyright{none}
\renewcommand\footnotetextcopyrightpermission[1]{}
\AtBeginDocument{%
  \providecommand\BibTeX{{%
    \normalfont B\kern-0.5em{\scshape i\kern-0.25em b}\kern-0.8em\TeX}}}

\newcommand{\slonn}{\textsc{slo-nn}}
\newcommand{\slonns}{\textsc{slo-nn}s}
\newcommand{\freehash}{\textsc{FreeHash}}
\usepackage{amsthm} %new
\usepackage{adjustbox} %new
\usepackage{algorithm}
\usepackage{algpseudocode}
\usepackage{hyperref}
\newtheorem{definition}{Definition} %new
\begin{document}

\title{Dynamic Network Adaptation at Inference}

\author{Daniel Mendoza and Caroline Trippel}
\affiliation{%
  \institution{Stanford University}
  %\streetaddress{P.O. Box 1212}
  \country{}
}
\email{{dmendo, trippel}@stanford.edu}

\begin{abstract}
Machine learning (ML) inference is a real-time workload that must comply with strict Service Level Objectives (SLOs), including latency and accuracy targets.
Unfortunately, ensuring that SLOs are not violated in inference-serving systems is challenging due to inherent model accuracy-latency tradeoffs, SLO diversity across and within application domains, evolution of SLOs over time, unpredictable query patterns, and co-location interference.
In this paper, we observe that neural networks exhibit high degrees of per-input activation sparsity during inference. . 
Thus, we propose \textit{SLO-Aware Neural Networks} (\slonns{}) which dynamically drop out nodes \textit{per-inference query}, thereby tuning the amount of computation performed, according to specified SLO optimization targets and machine utilization.
\slonns{} achieve average speedups of $1.3-56.7\times$ with little to no accuracy loss (less than 0.3\%).
When accuracy constrained, \slonns{} are able to serve a range of accuracy targets at low latency with the \textit{same trained model}. When latency constrained, \slonns{} can proactively alleviate latency degradation from co-location interference while maintaining high accuracy to meet latency constraints.
\end{abstract}


%\ccsdesc[500]{Computer systems organization}


\maketitle
\pagestyle{plain}

\section{Introduction}
\label{sec:intro}

Machine Learning (ML) inference supports many important application domains such as ranking and recommendation \cite{dlrm_oss}, finance \cite{Dixon2020}, analytics \cite{zhang:video, jiang:chameleon}, computer vision \cite{imagenet, resnet}, healthcare \cite{inference:healthcare}, computer security \cite{mireshghallah:shredder}, natural language processing \cite{devlin-etal-2019-bert}, and more. Thus, ML inference is at the heart of modern web services.
%, with companies such as IBM, Google, Amazon, and Microsoft offering Machine Learning as a Service (MLaaS) platforms \cite{...}.
% Machine Learning as a Service (MLaas) platforms: IBM Watson, Google Cloud Prediction, Amazon ML, Microsoft Azure ML.
For example, at Amazon Web Services (AWS), machine learning inference accounts for more than 90\% of infrastructure costs \cite{aws:inference}. At Facebook, more than 200 trillion predictions and over 6 billion languages translations are made each day \cite{facebook:inference}.

Unlike training which can be done offline, ML inference is a real-time workload that must comply with strict Service Level Objectives (SLOs), such as latency and accuracy targets. Unfortunately, ensuring that SLOs are not violated, complicates the design of inference-serving systems for a few key reasons outlined below.

\begin{figure}
    \centering
    \includegraphics[width=0.44\linewidth]{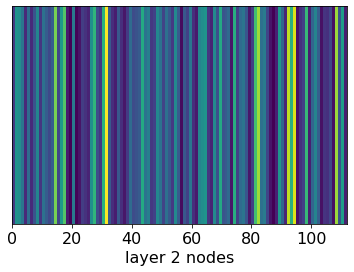}\includegraphics[width=0.48\linewidth]{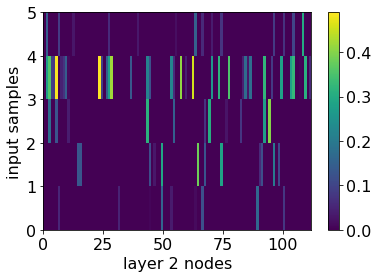}
    \caption{Left: Average per-node activation magnitudes for a 112-node hidden layer over 10,000 FMNIST input samples. Right: Per-node activations for five random inputs for the same hidden layer.
    % Bright colors indicate higher activation magnitudes.
    \slonns{} exploit the extreme sparsity of
    % the fact that
    per-node activations on behalf of individual inputs
    % are extremely sparse
    %per-input activations exhibit high degrees of sparsity
    compared to the average.
    %to reduce computation during neural network inference.
   % Note that per-input activations are significantly more spare he activation for each input is much more sparse than the average. We desire to exploit the per-input activation sparsity to reduce computation in the neural network.
    }
    \label{fig:sparse}
\end{figure}

\noindent \textbf{Latency-accuracy tradeoff:} First, highly accurate models often exhibit longer inference latencies than moderately accurate models, indicating a challenge in selecting a model which meets both accuracy and latency targets.

\noindent \textbf{Application SLO diversity:} Second, SLOs vary widely \textit{across} and \textit{within} application domains \cite{hazelwood:2018:mlatfb,romero:infaas}, requiring many \textit{model-variants} to satisfy diverse requirements. 
%For example, ML inference systems for healthcare and finance might prioritize accuracy above all else, while social networking and e-commerce services might prioritize responsiveness. 
%Or, within the domain of face recognition, some use-cases like intruder detection may prioritize low latency, whereas face tagging in social media applications may prioritize accuracy \cite{romero:infaas}.

\noindent \textbf{Evolution of SLOs over time:} Third, service providers may change SLOs over time. Suitable model-variants today may fail to satisfy SLOs in the future when combined with new compute infrastructure or deployed in a new execution environment \cite{SLO_google}.

\noindent \textbf{Co-location interference:}
Fourth, inference models are typically co-located on \textit{worker} machines to improve resource utilization and reduce operating costs \cite{lemay2020perseus,inter_inf,horus,romero:infaas}.
Unfortunately, model co-location introduces the opportunity for model interference, which can degrade inference latency and cause SLO violations.
% Unlike training which can be done offline, ML inference is a real-time workload that must comply with strict, Service Level Objectives (SLOs), such as latency and accuracy. Furthermore, SLOs vary widely with the application domain and use-case \cite{hazelwood:2018:mlatfb,romero:infaas, more}. ML inference systems for healthcare and finance might prioritize accuracy above all else, while social networking and e-commerce services might prioritize responsiveness.
% Each of these inference applications exhibit a diversity of Service Level Objectives (SLOs) \cite{aws_inferentia, deeprecsys} such as inference latency constraint and target accuracy. 

\noindent \textbf{Volatile query patterns:} Fifth, query arrival patterns are difficult to predict, and query rates fluctuate over time \cite{deeprecsys}. Query load impacts queuing times as well as intermittent co-location latency degradation. 

Given the challenges above, modern model serving systems~\cite{crankshaw2017clipper,romero:infaas} are burdened by training and managing many models to meet diverse SLOs under varying query loads, where switching models is prohibitively time consuming.
%As observed by \cite{lemay2020perseus}, model load times are often significantly longer (up to 100$\times$ slower) than inference; thus, switching models online is likely to lead to latency SLO violations.

This paper presents \textit{SLO-Aware Neural Networks} (\slonns{})---neural networks which are able to dynamically adapt inference computation on a \textit{per-input} basis to meet SLO optimization targets, even in the presence of co-location interference. Our insight, as suggested by prior work, is that per-input node activations in neural networks with ReLU activations are often sparse~\cite{adaptive_dropout,MONGOOSE,samplesoftmax,spring2016scalable}.
%The presence of node activation sparsity has been suggested by many prior work~\cite{adaptive_dropout,MONGOOSE,samplesoftmax,spring2016scalable}.
Thus, full-network inference accuracy can in theory be achieved at lower latency by using only a \textit{subset} of the network's nodes. Fig.~\ref{fig:sparse} illustrates this observation using a neuron pruned~\cite{neuronpruning,blalock2020state} model trained on the FMNIST data set \cite{xiao2017fashionmnist}. 
%The trained network was pruned \cite{neuronpruning,blalock2020state} to ensure the importance of each node for accurate inference. 
The figure shows that individual inputs exhibit extreme sparsity in the nodes they activate (left), despite node activations appearing dense when averaged across 10,000 input samples (right).
% Pre- and post inference accuracy on 
%\textbf{Our second insight} is that similar inputs exhibit similar per-node activations. 
%Thus, if one can identify a relevant subset of nodes for one input the same subset can be used to perform high-accuracy inference for another similar input.

\slonns{} leverage the above insight to optimize inference for SLOs by selectively dropping out nodes at inference time on a per-input basis, \textit{avoiding computations for these nodes altogether}.
% Fig.~\ref{fig:general_idea} illustrates this idea. 
Given a trained neural network (\slonns{} place no restrictions on model training or architecture), \slonns{} deploy a \textit{Node Activator} at each layer (see Fig.~\ref{fig:overall_LSH}) that dynamically selects which node activations to compute for a given inference query.
\slonn{} Node Activators \textit{learn} the relative importance of nodes for groups of similar inputs.
% which activations are most likely in the absence of dropout.
%Specifically, \slonn{} Node Activators leverage a statistical technique, called \textit{Locality Sensitive Hashing} (LSH) \cite{classic_LSH}, to cheaply detect similarity between different inputs samples.
%Similar inputs are directed via a hash function to the same entries of LSH tables, which contain
%. Each table entry contains information, acquired during an unsupervised training procedure, on the \textit{relative node importance} and \textit{confidence} information for inputs that map to said entries; 
%high confidence suggests that a set of inputs exhibit high degrees of node activation sparsity---in this paper, we consider such inputs to be ``easy''. 
% Intuitively, we can think of inputs which exhibit high node activation sparsity as "easy" inputs since we may omit many computations in the neural network without affecting the prediction result compared to "hard" inputs which comparatively activate more nodes in the neural network.
%LSH tables in \slonns{} use a custom hash function called, \freehash{} (\S\ref{sec:free-hash}), which leverages trained neural network weight vectors to identify similarities between input vectors in constant time.
% They then take into
Given node importance information along with SLO optimization targets, machine utilization information, and query data features, Node activators selectively drop out nodes from inference computations.
% determine which nodes to drop out.
% in order to determine which nodes can/cannot be dropped out.
In this way, \slonns{} can simultaneously serve a variety of SLOs with just a single model.
%With a single pre-trained model, they can simultaneously achieve the low latency of a small neural network model and the high accuracy of a large model.
We summarize our contributions as follows:

% The Node Activator is trained in an unsupervised manner to learn (1) similarity between different inputs, (2) relative activation importance for groups of similar inputs, and (3) confidence of the neural network for groups of similar inputs, where high confidence inputs are those which exhibit high degrees of activation sparsit

% \noindent \textbf{Dynamic dropout at inference:} To our knowledge, we propose first generic framework for dynamic dropout at inference that can be tuned according to accuracy and latency cost functions. This framework uses an LSH-based scheme to identify similarities between data inputs and leverages per-input activation sparsity to reduce the amount of computation needed to serve an inference request at a target accuracy.
% is SLO-aware and interference-aware.

\noindent \textbf{\slonns{} for SLO- and interference-aware inference:}
We propose \slonns{}, which to our knowledge, represent the first generic framework for dynamic dropout at inference with no restrictions on the model architecture or model training. 
%\slonns{} use an LSH-based scheme to identify similarities between data inputs and leverage per-input activation sparsity to tune the amount of computation needed to serve an inference request according to per-input SLO optimization targets and intermittent co-location interference.

%\noindent \textbf{\freehash{}:} To achieve high accuracy in classifying sets of inputs as \textit{similar} and thereby directing them to the same LSH table entries, we propose the \freehash{} hash function, which leverages trained neural network weights to derive input similarity.% in constant time.
% Algorithmic innovations designed to make LSH for dropout at inference fast and accurate

\noindent \textbf{\slonn{} case study:} We demonstrate the efficacy of \slonns{} across five neural network architectures and datasets: FMNIST~\cite{xiao2017fashionmnist}, FMA~\cite{fma_dataset}, 
%Kitsune~\cite{kitsune}, 
Wiki10~\cite{wiki10}, AmazonCat-13K~\cite{amazon13k}, and Delicious-200K~\cite{delicious200k}.
%The approach takes as input a pre-trained neural network and transforms it into a \slonn{}.
\slonns{} achieve average speedups ranging $1.3-56.7\times$ with zero or negligible accuracy difference (less than 0.3\%) compared to the original neural network.
%We also exemplify \slonns{} are capable of simultaneously serving a range of accuracy targets while minimizing latency, where lowering an accuracy target results in significant increased speedups.
%When accuracy constrained, \slonns{} are able to serve a range of accuracy targets at low latency with the \textit{same trained model}. 
%When latency constrained, \slonns{} can proactively alleviate latency degradation from co-location interference while maintaining high accuracy to meet latency constraints.
%In particular, we observed that the \slonn{} inference latency increases by $0-0.68\times$ for minimal loss in accuracy (less than 0.1\%) where the original neural network latency increases by $0.7-1.5\times$ due to co-location interference.

\section{SLO-Aware Neural Networks}
\label{sec:slo-dropout}

\begin{figure}[t]
    \centering
    \includegraphics[width=\linewidth]{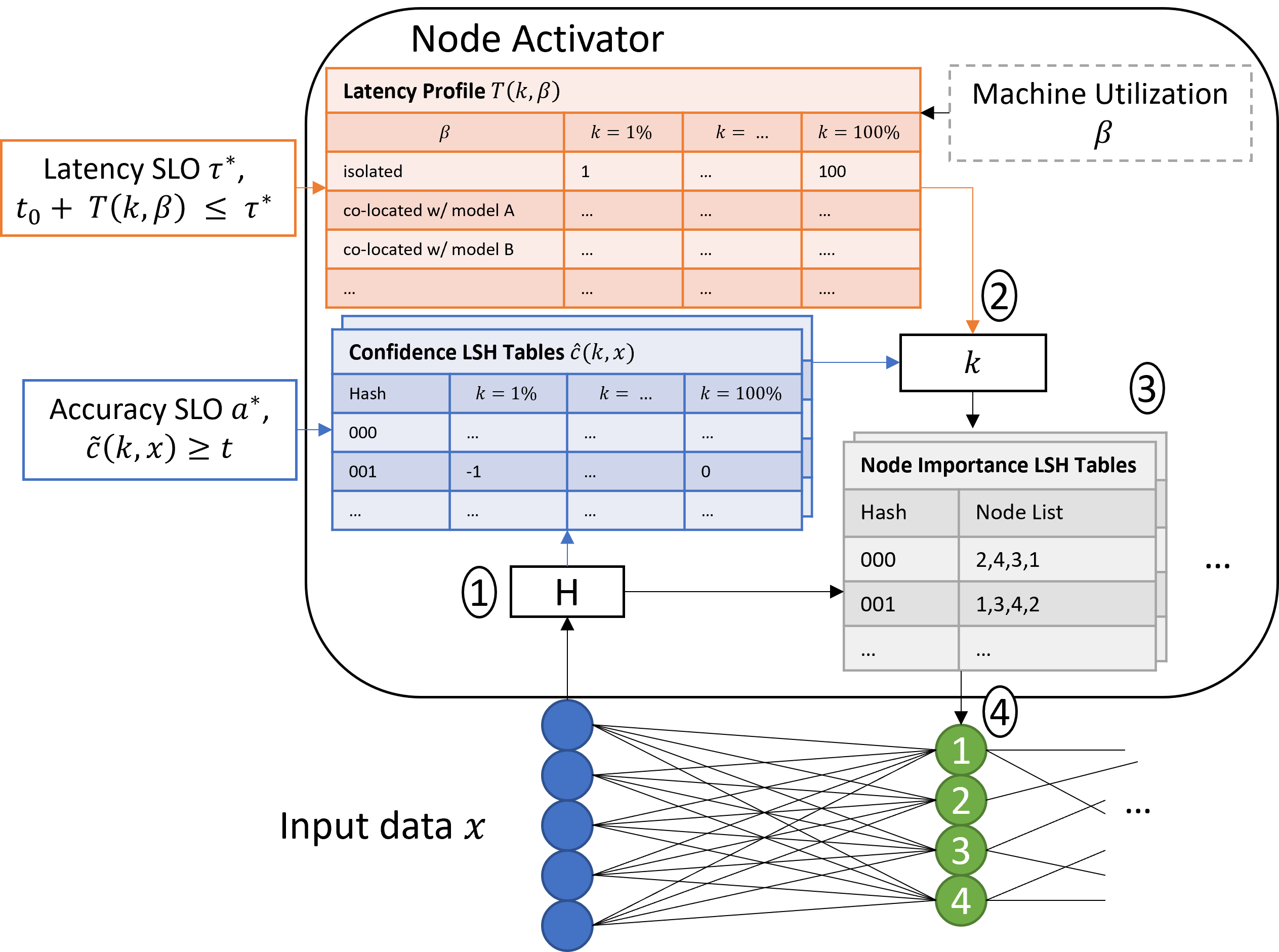}
    \caption{\slonn{} Architecture. (1) \slonn{} inputs are hashed (2) A particular percentage of nodes, $k$, is determined based on an SLO optimization target (e.g., ACLO or LCAO) and worker machine utilization information. 
    %With ACLO, the confidence LSH tables are used to determine confidence on the input data. With LCAO, the Latency Profile is used. 
    (3) Extract a list of nodes, sorted by importance from per-layer Node Importance LSH Tables.
    % The $k$\% number of nodes activated at the layer is determined depending on the SLOs and optimization objective.
    %For optimizing latency, the confidence LSH tables is used to minimize $k$ while ensuring the neural network will meet the accuracy target. For optimizing utilization, the latency profile $T(k,\beta)$ is used to maximize latency while still meeting the latency SLO.
    % 3) The Node list LSH is queried to obtain a node list sorted by importance.
    (4) The top $k$\% nodes in the sorted list are computed per layer.
    % Notice that after the first layer, step 2 is skipped as $k$ is determined only at the first layer.
    }
    \label{fig:overall_LSH}
\end{figure}

Fig.~\ref{fig:overall_LSH} illustrates the \slonn{} architecture. 
\slonns{} dynamically optimize neural network inference computations on a per-query basis given (1) an SLO optimization target and (2) information about the machine utilization of the worker machine on which it is running. Our implementation of \slonns{} supports two SLO optimization targets: \textit{Accuracy-Constrained Latency-Optimized} (ACLO) and \textit{Latency-Constrained Accuracy-Optimized} (LCAO). ACLO can be used to maximize query throughput and minimize co-location interference on behalf of a particular model. On the other hand, LCAO can leverage information about current machine utilization, to adapt to changing co-location interference or query loads without violating latency SLOs.

\subsection{General Framework}
\label{sec:framework}
In this section, we present some terminology which we use to define \textit{SLO-Aware Neural Networks} and further describe the mechanics of the ACLO and LCAO optimization targets.
% We now introduce some notation that will be used to describe the operation of \slonns{}.
To begin with, an inference \textit{query} consists of: (1) an accuracy target $a^*$, (2) a latency target $\tau^*$, and (3) input features $x$. For each query, \slonns{} dynamically tune computation according to the network's \textit{confidence} as well as supplied \textit{accuracy} and \textit{latency} constraints. We describe these three parameters as follows.

\textbf{Confidence:} Let $c(k,x)$ represent a neural network's confidence when performing inference on data input $x$ with the top $k\%$ of nodes \textit{at each layer} computed (not dropped out). In \slonns{}, the top $k\%$ nodes are selected with respect to per-layer lists of nodes that are ranked by importance, where importance corresponds to the expected activation magnitude.
\S\ref{sec:node-importance} describes how these ranked lists are constructed.
% In \slonns{}, precisely which nodes comprise the $k\%$ that are computed is determined by the Node Activator, which is described in \S\ref{sec:free-hash}.
For a given input, we quantify confidence as the negative distance between the prediction of the full neural network, $\hat{y}$, and the prediction of the neural network with the top $k\%$ of nodes computed, $\hat{y}_k$.
\begin{equation}
    c(k,x) = -distance(\hat{y},\hat{y}_k)
\end{equation}
The $distance$ function for computing $c(k,x)$ is selected based on the prediction task.
%of the neural network.
For instance, we chose the $distance$ function to be cross-entropy for classification tasks.

\textbf{Accuracy:} Let $a_t$ be the measured accuracy on a held-out set where a neural network predicted every data input $x$ with confidence $c(k,x) \geq t$; $t$ is some confidence threshold.

\textbf{Latency:} Let $T(k,\beta)$ denote the latency of the neural network when $k\%$ of the nodes are activated given the state of the execution environment $\beta$. 
%$\beta$ represents parts of the execution platform which may influence a model's latency, including the current machine utilization and query load.
$\beta$ represents the machine utilization on behalf of co-located workloads which may cause interference and increase inference latency.
%$\beta$ represents parts of the execution platform which may influence the model's latency, including machine utilization and co-located workloads which may contend for resources and degrade the inference latency of the neural network.
Further, let $t_0$ be the total time spent processing the query outside of inference including queuing delays and feature extraction.
Thus $t_0 + T(k,\beta)$ denotes the total time spent processing the query.
Note that $t_0$ may vary from query to query (e.g. due to varying queuing delays).

\begin{definition}[SLO-Aware Neural Network]
Given accuracy target $a^*$, latency constraint $\tau^*$, an \slonn{} chooses $k$ such that: $a_{c(k,x)} \geq a^*$ and $t_0 + T(k,\beta) \leq \tau^*$.
% \begin{itemize}
%     \item $a_{c(k,x)} \geq a^*$
%     \item $T(k,\beta) \leq \tau^*$
% \end{itemize}
If these constraints cannot be met, then the neural network cannot fulfill the SLOs.
\end{definition}
Note that there may be a range of $k$ which may satisfy the SLOs.
This range represents the degree of freedom for which the \slonn{} can adapt to each query.
In this paper, we consider \slonns{} which optimize for one SLO (accuracy or latency), while constrained by another---the ACLO and LCAO optimization targets.

%Algorithm \ref{alg:predict} demonstrates the forward propagation of an \slonn{} with $k\%$ activation at each layer. The Node Activator provides a list of the $k$\% nodes to activate in the current layer.

%\begin{algorithm}
%\caption{$k\%$ Forward Propagation}\label{alg:predict}
%\textbf{Input:} $x$, $k$ \\
%\textbf{Output:} Activation
%\begin{algorithmic}[1]
%\STATE input := $x$
%\FOR{$l = 1$ : Layers}
%  \STATE ActiveNodeIDs := NodeActivator.getTopK($k$,input)
%  \STATE Activation := ForwardProp(input,ActiveNodeIDs)
%  \STATE input := Activation
%\ENDFOR
%\end{algorithmic}
%\end{algorithm}

\subsection{Accuracy-Constrained Latency-Optimized}
\label{sec:aclo}
\slonns{} can be used to minimize inference latency
%We demonstrate that the \slonn{} can be leveraged to minimize latency
$T(k,\beta)$ for each input query while satisfying an accuracy target $a^*$.
$t_0 + T(k,\beta)$ monotonically decreases with $k$ when $t_0$ is held constant since decreasing $k$ can only decrease or not affect $T(k,\beta)$. Therefore, minimizing latency $t_0 + T(k,\beta)$ is equivalent to minimizing $k$.
%As stated previously, a accuracy target $a_t$ corresponds to a confidence threshold $t$ such that every data point with predicted with confidence $c(k,x) \geq t$
Thus the optimization problem, corresponding to our ACLO SLO optimization target, is expressed as follows:
\begin{equation}
\begin{array}{cl}
\displaystyle \min_{k} & k\\
\textrm{s.t.} & a_{c(k,x)} \geq a^*\\
\end{array}
\end{equation}
Note that confidence depends on
% upon
$x$, indicating that for ``easy'' inputs
%where the neural network exhibits high confidence, 
\slonns{} can drop out significantly more computation. Thus, a large high-accuracy model can adapt to serve a range of accuracy constraints where more lenient accuracy targets likely correspond to significantly lower inference time.
%Furthermore, by minimizing inference latency at a given accuracy target, the ACLO scheme can maximize inference throughput and minimize compute footprint, thereby reducing a neural networks co-location interference with respect to other workloads. 
% Minimize the number of activated nodes at this accuracy target
% + Lowers inference latency at a given accuracy target --> increased throughput
% + Lowers the compute footprint --> lower co-location interference
% \begin{itemize}
%     \item Dynamically adapt neural network for each input query to minimize inference latency
%     \item One model able to serve a range of accuracy constraints where more lenient accuracy targets likely correspond to significantly lower inference time
%     \item Minimizing latency in turn reduces the interference the neural network causes to other co-located workloads
% \end{itemize}

%Algorithm \ref{alg:maxdrop_predict} shows the strategy to minimize latency of the inference for each query.

%\begin{algorithm}
%\caption{Minimal Latency Prediction}\label{alg:maxdrop_predict}
%\textbf{Input:} $x$, $a^*$ \\
%\textbf{Output:} $\hat{y}_k$
%\begin{algorithmic}[1]
%\STATE $k$ := min(\{$k'$ $|$ $c(k',x) \geq t \land a_t \geq a^*$ \})
%\STATE $\hat{y}_k := \textsc{k\%ForwardPropagation}(x,k)$
%\end{algorithmic}
%\end{algorithm}

\subsection{Latency-Constrained Accuracy-Optimized}
\label{sec:lcao}
\slonns{} can also be used to optimize inference accuracy $a_{c(k,x)}$ per input query while satisfying a latency target $\tau^*$.
In all of our experiments (\S\ref{sec:results}), we observe that as $k$ increases, $a_{c(k,x)}$ either monotonically increases or approaches the accuracy of the full neural network.
% (for all evaluated models and datasets). 
%Note that $a_{c(k,x)}$ with $k=100$ corresponds to the accuracy of the neural network with a 100\% of the nodes computed. 
%For the one outlier experiment, we observed better-than-full-network accuracy when only a small fraction of the model was computed ($k<1\%$) per input. 
%Further increasing $k$ resulted in an accuracy decrease which approached full-network accuracy. 
%Therefore, our interpretation of ``optimizing accuracy'' is minimizing the distance between $c(k,x)$ and $c(100,x)$.

Since the percentage of computed nodes, $k$, is an indicator of both latency (lower $k$ implies lower latency) and accuracy (higher $k$ implies accuracy closer to the full neural network), we can leverage a cost function based on $k$ to optimize inference accuracy. Namely, we can maximize $k$, and thus inference accuracy $a_{c(k,x)}$, such that the latency constraint $\tau^*$ is satisfied just-in-time.
Therefore, the optimization problem, corresponding to our LCAO optimization target, is expressed as follows:
\begin{equation}
\begin{array}{cl}
\displaystyle \max_{k} & k\\
\textrm{s.t.} & t_0 + T(k,\beta) \leq \tau^*\\
\end{array}
\end{equation}   

%Notice that $k$ is selected independent from the accuracy constraint $a^*$, since we assume that accuracy maximizes when maximizing $k$.
%Supporting LCAO requires
%obtaining the latency profile $T(k,\beta)$ through either profiling or prediction. Previous studies have demonstrated how to accurate predict neural network latency given an exponentially sized parameter space~\cite{inter_inf}.
% The implementation of this approach involves
%profiling inference latency $T(k,\beta)$ for varying $k$ and varying co-location interference to anticipate the inference latency $T(k,\beta)$ at query time.
%We discuss our profiling strategy in \S\ref{sec:free-hash}.
%Specifically, profiling enables \slonns{} to associate a given $k$ with a particular latency when supplied information about the machine utilization of the worker machine on which it is running. 
%In our experiments, we consider two execution scenarios when profiling---one in which a given neural networks serves inference queries in isolation and another in which it is co-located with an identical model.
%Note that latency profiling is not necessary for ACLO, since ACLO is accuracy constrained and minimizes $k$ such that the accuracy target is met. 
%Since each $k$ is associated with some accuracy for a given class of inputs (those which map to the same entry of an LSH table), ACLO need only pick the minimum $k$ needed to attain a particular accuracy requirement.

One benefit of \slonns{} which deploy the LCAO optimization is that they are able to serve a wide range of latency SLO targets with a single model. Furthermore, LCAO \slonns{} can adapt to intermittent co-location interference or bursty and varying query loads by using a latency profile to predict inference latency and dynamically adjusting neural network compute accordingly. In doing so, LCAO \slonns{} can avoid latency SLO violations where a standard neural network would not be able to satisfy.

\begin{table}[t!]
\begin{center}
\begin{adjustbox}{width=1\linewidth}
\begin{tabular}{ |c|c|c|c|c|c| }
 \hline
 Dataset & Train size & Test size & Feature dim & Label dim & Architecture \\ 
 \hline
 FMNIST & 60,000 & 10,000 & 782 & 10 & 112-112\\
 FMA & 84,353 & 22,221 & 518 & 161 & 64\\
 Wiki10 & 14,146 & 6,616 & 101,938 & 30,938 & 128\\
 AmazonCat13k & 1,186,239 & 306,782 & 203,883 & 13,330 & 128\\
 Delicious200k & 196,606 & 100,095 & 782,585 & 196,606 & 128\\
 \hline
\end{tabular}
\end{adjustbox}
\end{center}
\caption{\slonn{}-evaluated datasets and model architectures.}
\label{tab:datasets}
\end{table}

\section{\slonn{} Node Activators}
\label{sec:node-activator}
In this section, we describe \slonn{} \textit{Node Activators}, which select nodes to be dropped out for a given inference request and SLO optimization target (ALCO versus LCAO). In this work, the Node Activator is based on Locality Sensitive hashing (LSH) due to its low overhead. In future work we plan to investigate other ranking schemes.
% In this section, we describe how we designed the LSH-based Node Activator for \slonn{}s and introduce a hash scheme for \slonn{}s called free hash.
% Our approach minimizes the overhead of the LSH at inference time while achieving high accuracy.

\subsection{Locality Sensitive Hashing}
\label{sec:lsh}
LSH was originally proposed as a sub-linear time approximate nearest neighbors search strategy \cite{classic_LSH}. The technique features a family of hash functions with the property that similar input objects have a higher probability of colliding (post-hash) than non-similar ones given some similarity measure.
% LSH is often employed as sub-linear time complexity solution for approximate nearest-neighbor search~\cite{classic_LSH}. 
In particular, a sufficient condition for a family of hash functions $\mathcal{H}$ to be considered an \textit{LSH family} is that for $h\in\mathcal{H}$, the post-hash collision probability $Pr_{\mathcal{H}}(h(x) = h(y))$ monotonically increases with the similarity of $x$ and $y$. 
%Popular LSH families such as \textsc{SimHash}~\cite{SimHash} and \textsc{WTAHash}~\cite{WTAHash} have been shown to adhere to this property. 
%More precisely, where $f$ is a monotonically increasing function:
%\begin{equation}
%    Pr_{\mathcal{H}}(h(x) = h(y)) = f(Sim(x,y))
%\end{equation}

% where $h$ is chosen from $\mathcal{H}$ \cite{MONGOOSE}. 

%Formally, Let $H$ be a family of functions which map $R^d$ to some  $S$.
%\begin{definition}[LSH Family]
%A family $H$ is called $(S_0,cS_0,p_1,p_2)$-sensitive if for any two points $x,y \in R^d$ and $h$ chosen uniformly from $H$ satisfies the following:
%\begin{itemize}
%    \item if $Sim(x,y) \geq S_0$ then $Pr(h(x) = h(y)) \geq p_1$
%    \item if $Sim(x,y) \leq cS_0$ then $Pr(h(x) = h(y)) \leq p_2$
%\end{itemize}
%\end{definition}
%Where $c < 1$ and $p_1 > p_2$ are desirable properties in the LSH family for approximate nearest-neighbor search. In essence, an LSH is a probabilistic approach to the nearest-neighbor search problem where selected hash functions of the LSH family are associated with a similarity measure. The primary advantage of applying LSH to approximate nearest-neighbor search is that a set of nearby neighbors according to the LSH family can be queried in sub-linear time complexity.

The classic $(K,L)$ LSH algorithm has two phases~\cite{classic_LSH}. In the \textit{pre-processing phase}, $L$ hash tables are constructed. For a given table, keys are computed by concatenating the outputs of $K$ LSH hash functions. 
%Different tables use different sets of $K$ hash functions to form keys. 
Data elements are then stored into buckets of the $L$ hash tables according to their computed keys.
%Fig.~\ref{fig:classic_LSH} illustrates LSH for $K=2$ and $L=1$. 
%Two hash functions $h_1$ and $h_2$ partition the data points into four regions. 
%Notice that nearby data points in space are mapped to the same bucket in the hash table by a two-hash function key.
In the \textit{query phase}, given some input query, keys for each hash table are computed and used to fetch all data elements from each of the $L$ corresponding buckets (one bucket per table). 
%The union of these bucket contents constitutes a set of similar data elements. 
\slonns{} leverage LSH to efficiently identify similar data inputs and further associate with them \textit{node importance} and \textit{confidence} information.

\subsection{Node Activator Training}
%\slonns{} apply \textit{Locality Sensitive Hashing} (LSH) to neural network inference. 
%Per-layer Node Activators (see Fig.~\ref{fig:overall_LSH}) leverage LSH to capture associations between similar inputs and node importance and confidence information. 
%For a given input, coupled with an SLO optimization target and machine utilization information, Node Activators dynamically drop out nodes at inference time, avoiding associated computations entirely.
% employs an LSH-based strategy which learns the association between input data and node activation. In \S\ref{sec:free-hash}, we detail our LSH-based dropout at inference approach. 

LSH provides a low overhead mechanism for associating similar data samples with each other---similar inputs collide in LSH hash tables.
\slonns{} further require associating each group of similar inputs with (1) a ranked list of nodes according to their importance for making accurate predictions (i.e., as close to the full neural network as possible) and (2) a confidence score which encodes their ``hardness''. 
As illustrated in Fig.~\ref{fig:overall_LSH}, the Node Activator leverages two types of LSH hash tables for storing each association type---the \textit{Node Importance} (gray) and \textit{Confidence} (blue) tables, respectively. 
%Note that only one set of Confidence tables are required for the full network, whereas a set of Node Importance tables are associated with each non-input layer.
The hash tables which make up the Node Activator are populated with the help of an unsupervised training step, which can be performed pre- or post-deployment of the \slonn{}. 
% trained in an unsupervised manner to learn these two associations.
% training step where an input dataset is used to learn associations between node activation and input data.
% Since training is unsupervised, it can be performed pre- or post-deployment of the \slonn{}. 
%Furthermore, Node Activators can be periodically trained post-deployment on new data received from input queries.

\noindent \textbf{Node Importance LSH Tables}
\label{sec:node-importance}
In an \slonn{}, a set of Node Importance LSH tables (gray tables in Fig.~\ref{fig:overall_LSH}) are placed at each layer. Node Importance tables map a set of similar inputs (which collide in the same table entry) to a list of nodes, ranked according to their importance in facilitating accurate inference. 
% importance in supporting accurate predictions for said set of imputs.
%The Node Activator contains an LSH which contains a mapping between input data and a node list sorted by node importance. 
For a given input, ranking nodes in a specific layer according to their importance corresponds to ranking them according to their activation magnitude.
% by To quantify node importance for a given input, we use the magnitude of the node activation.

Algorithm \ref{alg:LSH dropout} describes the unsupervised training procedure for a set of $L$ Node Importance LSH tables at some layer $l$ in a \slonn{}. 
% the pre-processing phase to build the LSH for the Node Activator at layer $l$.
% The training phase takes as
Inputs to the training procedure include an
input set of data features, $Input_{l}$, and LSH parameters, $L$ (number of tables) and $K$ (key size).
The dataset $Input_{l}$ is representative of the data which is supplied as input to layer $l$ of the neural network during inference.
% The procedure returns $L$ trained the hash tables and their corresponding hash functions.
%Algorithm \ref{alg:LSH dropout} first generates $K \times L$ hash functions (i.e, $K$ hash functions per table), according to some LSH hash family of choice (\S\ref{sec:free-hash}).
% We discuss hash function generation in \S\ref{sec:free-hash}.
Training is initialized by first generating $K \times L$ hash functions (i.e, $K$ hash functions per table), according to some LSH hash family of choice (\S\ref{sec:free-hash}).
Next, for each of the $L$ tables, the algorithm computes the $K$ corresponding hash functions over all inputs $x \in Input_l$, thereby mapping each input $x$ to a particular bucket in each table. 
For inputs which map to the same bucket in some table, their per-node activations (at layer $l$ where the table is positioned) are summed; the result is an activation sum associated with each node.
% and sums the activations which are hashed to the same bucket.
Finally, nodes are sorted according to activation sums (highest to lowest). 
%The result is stored as a list of node IDs in the appropriate bucket.
% each of the buckets are sorted and the list of node IDs sorted by importance is stored.

% \begin{figure*}
%     \centering
%     \includegraphics[width=\linewidth]{figures/paper_figs/overall_LSH.png}
%     \caption{Architecture of LSH-based Node Activator. 1) the hash functions are computed for the LSH tables. 2) The $k$\% number of nodes activated at the layer is determined depending on the SLOs and optimization objective. For optimizing latency, the confidence LSH tables is used to minimize $k$ while ensuring the neural network will meet the accuracy target. For optimizing utilization, the latency profile $T(k,\beta)$ is used to maximize latency while still meeting the latency SLO. 3) The Node list LSH is queried to obtain a node list sorted by importance. 4) The top $k$\% nodes in the node list are computed for the layer. Notice that after the first layer, step 2 is skipped as $k$ is determined only at the first layer.}
%     \label{fig:overall_LSH}
% \end{figure*}

\begin{algorithm}
\caption{Training Node List LSH at layer $l$}\label{alg:LSH dropout}
\textbf{Input:} $Input_{l}$, $L$, $K$ \\
\textbf{Output:} LSH
\begin{algorithmic}[1]
\State $h$ := GenerateHashFamily($K$,$L$)
\State NodeScore := Map
\State Rank := Map
\For{$x \in Input_{l}$}
  \For{b = 1 : $L$}
    \State key := $(h^b_1(x),h^b_2(x),...,h^b_K(x))$
    \State Activation = ForwardProp$_l$($x$)
    \State NodeScore[b][key] += Activation
  \EndFor
\EndFor
\For{b = 1 : $L$}
  \For{key $\in$ NodeScore[b]}
    \State Rank[b][key] := argsort(NodeScore[b][key])
  \EndFor
\EndFor
\State LSH := (Rank,$h$)
\end{algorithmic}
\end{algorithm}

\noindent \textbf{Confidence LSH Tables}
\label{sec:node-confidence}
% \noindent \textbf{Confidence Estimation}
As discussed in \S\ref{sec:framework}, $c(k,x)$ is a measure of confidence the neural network exhibits on data input $x$ when the top $k\%$ of nodes are activated at each layer.
%Given some accuracy target $a^*$, the neural network must activate $k$\% nodes such that $c(k,x) \geq t$ and $a_t \geq a^*$.
%A naive approach to calculate confidence would be to directly calculate $c(k,x)$ which requires computing the predictions of the full neural network and the neural network with $k\%$ node activations, which defeats the purpose of dropout at inference to reduce computation.
% Our insight to get around this problem is to
%employ LSH to estimate confidence.
Intuitively, a given neural network will exhibit a similar level of prediction confidence when supplied with similar input features.
Thus, \slonns{} leverage a set of Confidence LSH tables (blue tables in Fig.~\ref{fig:overall_LSH}) to associate groups of similar inputs with a confidence score.
%Thus, we create a confidence LSH which uses the same hash functions of the node list LSH.
%The blue hash tables in Fig.~\ref{fig:overall_LSH} demonstrate the LSH for computing confidence on each input.

Let $\hat{c}(k,x)$ be an estimate of confidence $c(k,x)$ such that
\begin{equation}
    \hat{c}(k,x) = aggregation(\{c(k,x') | x' \in LSH(x)\})
\end{equation}
Where $aggregation$ is a function which aggregates the confidences $c(k,x')$ of the data inputs that are hashed to the same bucket during the training procedure.
In our evaluation (\S\ref{sec:results}), the $aggregation$ function is the arithmetic mean, which relies on the intuition that nearby data inputs are likely to exhibit similar confidence on average.
%\slonns{} feature one set of $L$ Confidence LSH tables for the entire neural network (rather than per-layer as in the Node Importance tables of \S\ref{sec:node-importance}). Each entry in a Confidence LSH table associates a set of similar inputs with a confidence estimate, $\hat{c}(k,x)$, acquired from the training procedure.
% Note that $aggregation$ is computed at the pre-processing step of the LSH.
To associate a confidence threshold $t$ with an accuracy metric $a_t$, we test on the held-out validation set where we predict every data point $x$ with confidence $\hat{c}(k,x) \geq t$.

\noindent \textbf{Interference-Aware Latency Estimation}
%Anticipating the inference latency, $T(k,\beta)$, of an \slonn{} requires profiling or predicting latency on a particular worker machine.
%As stated in \ref{sec:lcao}, the \slonn{} selects $k$ so that $t_0 + T(k,\beta) \leq \tau^*$, where $t_0$ is query processing time outside of inference (such as queuing delays) and $\tau^*$ is the latency SLO.
% , we must profile latency on the worker machine.
For a given \slonn{}, inference latency $T(k,\beta)$ may be profiled or predicted apriori for varying co-location scenarios $\beta$ and varying values of $k$.
In our current experiments (\S\ref{sec:results}), we leverage latency profiles while in future work we plan to additionally train latency predictors which can be subsequently used 
%For latency prediction, we train a latency predictor which is then used
to predict a full latency co-location profile for a given workload configuration~\cite{quazar,inter_inf}.
%of the co-location configurations

%In other words, profiling considers $k=k_0$ and  $k=k_i$, for all $i$ such that $k_i\geq k_0$, $k_i\leq100$, and $k_{i+1}=k_i + \Delta$,. 
%Since $k$ expresses a node percentage, the number of nodes activated for each profiled value of $k$ is determined by rounding to the nearest whole node.
%In our evaluation, we set $k_0$ such that only one node is active in the network, and set increment interval $\Delta = 5$.

%Different values for $\beta$ correspond to different co-location scenarios (e.g. where the \slonn{} is co-located there models in an inference serving system). 
%One co-location scenario is possible for the case in which an \slonn{} executions in isolation ($\beta = isolated$); $n$ additional co-location scenarios are possible to represent various interfered execution possibilities ($\beta = co$-$located_n$).
%Different interfered executions correspond to different co-located workloads.
%The orange table in Fig.~\ref{fig:overall_LSH}, illustrates collected latency profiles for a particular \slonn{} which each contain latency measurements for various values of $k$ and $\beta$.
%profiling will produce 100 latency estimates corresponding to each possible valuation of $k$.
The Node Activator uses the estimated latencies when serving the model to anticipate the inference latency associated with a particular degree of dropout (i.e., a particular value of $k$) and co-location interference.
%This allows the \slonn{} to anticipate inference latency ($T(k,\beta)$) and drop out computation to alleviate co-location latency degradation or other processing delays $t_0$.
%so that it varies the amount of dropout to meet the latency deadline under co-location interference and bursty query load.
% The orange table in Fig.~\ref{fig:overall_LSH} illustrates latency profile $T(k,\beta)$

% We profile both in isolation and when co-located with other workloads (e.g. other models in the inference serving system) which may be deployed concurrently during the \slonn{}'s inference execution.

% with varying levels of $k$ when its co-located with other workloads such as other inference models.

% )$ of the \slonn{} is used to choose $k$.

%\begin{figure*}
%    \centering
%    \includegraphics[width=0.7\linewidth]{figures/paper_figs/latencymin_LSH.png}
%    \caption{LSH for latency minimizing framework}
%    \label{fig:latency_min_frame}
%\end{figure*}

%\begin{figure*}
%    \centering
%    \includegraphics[width=0.7\linewidth]{figures/paper_figs/utilmax_LSH.png}
%    \caption{LSH for utilization maximizing framework}
%    \label{fig:my_label}
%\end{figure*}

\subsection{SLO-Aware Forward Pass}
Fig.~\ref{fig:overall_LSH} illustrates the forward pass of \slonns{}.
% 1) At the input layer, first the hash functions are computed for the LSH tables. 2) The $k$\% number of nodes activated at the layer is determined depending on the SLOs and optimization objective. For optimizing latency, the confidence LSH tables is used to minimize $k$ while ensuring the neural network will meet the accuracy target. For optimizing utilization, the latency profile $T(k,\beta)$ is used to select $k$ in order to maximize latency such that the query meets its deadline. 3) The Node list LSH is queried to obtain a node list sorted by importance. 4) The top $k$\% nodes in the node list are computed for the layer. Notice that after the first layer, step 2 is skipped as $k$ is determined only at the first layer.
% \noindent \textbf{LSH Query}
Node Confidence LSH tables and the Latency Profile table are queried \textit{once per inference request} to select the percentage of nodes, $k$, to activate for a given SLO optimization target.
For ACLO, only the Node Confidence LSH tables are queried; for LCAO, only the Latency Profile table is accessed.
Node Importance LSH tables are queried \textit{once per layer} to obtain sorted lists of nodes
%according to their importance
from which the top $k$ can be selected.
\subsection{FreeHash: A Novel LSH Hash Function}
\label{sec:free-hash}
%To simultaneously address the issue of overhead and similarity of node activation exhibited by traditional LSH hashing approaches, we introduce Free Hash.
We observe that deploying LSH in the context of neural networks gives us access to a unique LSH hash family \textit{for free}. 
Specifically, neural network weights represent vectors that have been \textit{trained} to preserve the similarity between data inputs. 
Thus, 
%rather than computing dot products between input data and random vectors to deduce input similarity (as \textsc{SimHash} does), 
\slonns{} derive hash keys by computing dot products between input data and a sub-sample of the neural network weights.
%of the neural network.
%Specifically, at each layer, the weights and biases of $K\times L$ nodes are selected for computing hash keys.
%In other words, hash keys are computed from forward pass computations.
%The computation for the hash value is then the same computation done for the node during forward propagation. Thus, by computing the hash values, we are also doing forward propagation, and we re-use the hash values as outputs of the layer, hence the hash is \textit{free}.
We call this LSH hash family \freehash{}; for a given \slonn{} layer $l$, we define \freehash{} as follows.
%use the weights and biases of neural network layers.
% Free hash characterizes the similarity of data inputs in terms of their node activations while avoiding the overhead of conventional hashing methods.
% %Further we demonstrate in the results that \slonn{}s using Free Hash exhibit higher prediction accuracy than a popular cosine similarity hash called SimHash~\cite{SimHash}. 
% Free Hash is based on the insight that the weights of the neural network are trained to preserve the similarity of node activations between different data inputs.
% Thus, instead of random vectors, we may use the weights and biases of the layers in the neural network as the LSH hashing function.
% Formally, for a LSH at layer $l$ in the neural network, we define free hash as:
\begin{definition}[\freehash{}] Let $w_i$ and $b_i$ correspond to the weights and bias of some randomly selected node $i$ in layer $l$ of the \slonn{}. We hash an input $x$ to layer $l$ as:
\begin{equation}
    FreeHash_i(x) = sign(w_i^Tx + b_i)
\end{equation}
\end{definition}
%The result is a vector of sign bits. 
For ReLU Layers, free hash satisfies the LSH family hash condition of \S\ref{sec:lsh}.
% for the hamming distances between sets of active nodes:
%\begin{equation}
%    Pr(h(x)=h(y)) = f(-HammingDist(AS(x),AS(y)))
%\end{equation}
%where $f$ is a monotically increasing function and $AS$ is a function which returns the set of active nodes in the layer of the neural network.
%The above  property also holds for sigmoid activation function where the distance measure is the hamming distance between the set of activations greater than 0.5.
% Thus free hash exhibits a desirable property as inputs which have similar activations are likely to collide in the LSH.

%\noindent \textbf{Biased \freehash{}}
When using \freehash{} to construct hash functions for an \slonn{} LSH table, a set of $K\times L$ ($K$ keys per $L$ tables) nodes (and their corresponding weights and biases) must be selected. 
%Note that nodes which are selected for \freehash{} are not selected by the Node List LSH since the \freehash{} always computes those nodes. 
%(and thus they do not need to be explicitly selected in the Node List LSH).
%For Confidence LSH Tables, the nodes selected come from the input layer. For Node Importance LSH tables, the nodes selected come from the tables' associated neural network layer. 
Theoretically, these nodes could be selected at random from the relevant layer. However, this approach may result in disimilar data inputs, which produce sparse activations for a given neural network layer, being misclassified as similar.
%may be similar in terms of hamming distance, but do not activate the same nodes.
%make free hash more amenable to sparse activation, we
To address this issue, \slonns{} sample node weights and biases for \freehash{} with probability proportional to the variance of the nodes' activations across the training set for the LSH. 
%More formally,
%In other words, assuming $h$ is the set of \freehash{} functions (i.e., node weight an bias vectors) for the LSH sampled from the set of all \freehash{} functions, the probability that $FreeHash_i\in h$ is:
%\begin{equation}
%    Pr(FreeHash_i \in h) \propto Var(Activation_i)
%\end{equation}
\freehash{} leverages computations that are already required to perform full neural network inference. Thus, in the worst case, where all nodes in an \slonn{} are computed, no extra computation is required compared to the full neural network.
% \noindent \textbf{Storage Overhead}
Furthermore, the Node Activator is a lightweight data structure as it stores sparse tables with lists of node references. In our evaluation, Node Activator storage accounted for less than 10\% of the neural network for all models.

\section{Methodology}
\label{sec:method}
\begin{figure*}[t]
\centering 
\includegraphics[width=0.19\linewidth]{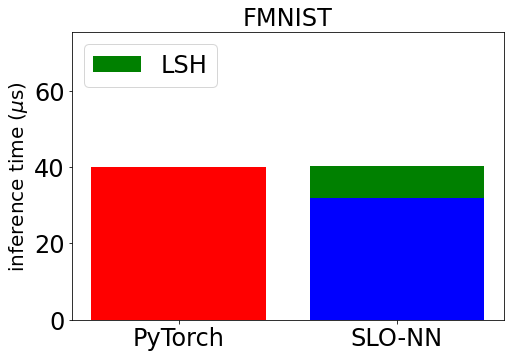}
\includegraphics[width=0.19\linewidth]{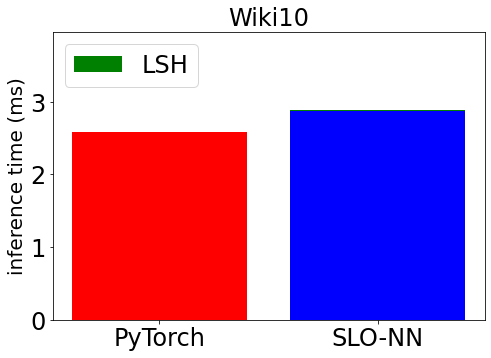}
\includegraphics[width=0.19\linewidth]{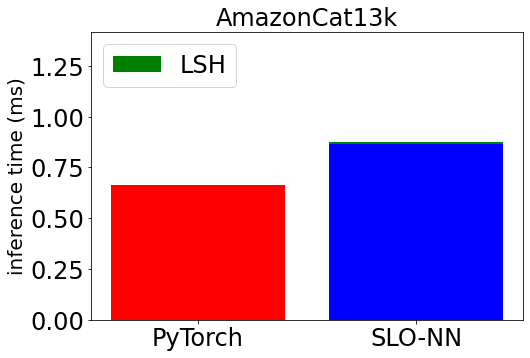}
\includegraphics[width=0.19\linewidth]{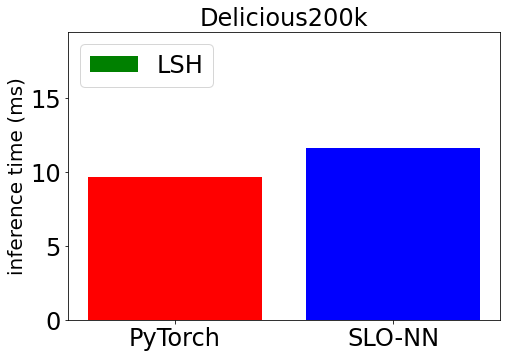}
\includegraphics[width=0.19\linewidth]{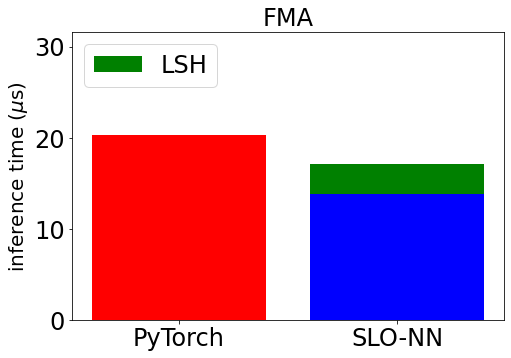}
\\
\caption{
Performance breakdown of full neural network inference latency (all nodes computed) for a baseline \textit{PyTorch} implementation and an \slonn{}. The \slonn{} is using \freehash{}. The time spent in the LSH includes the time spent computing hash functions. Our framework performs similarly to PyTorch, demonstrating the practicality of our \slonn{} approach.
}
\label{fig:latency_breakdown}
\end{figure*}

\begin{figure*}
\centering 
\includegraphics[width=0.19\linewidth]{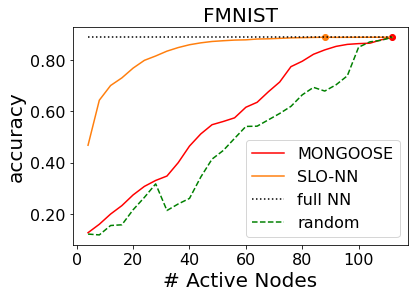}
\includegraphics[width=0.19\linewidth]{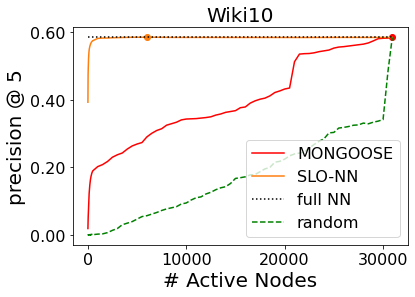}
\includegraphics[width=0.19\linewidth]{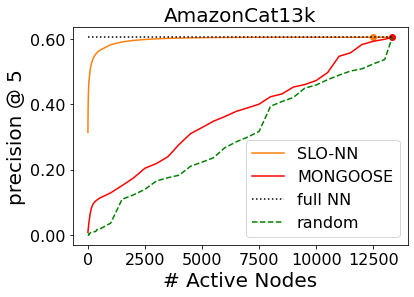}
\includegraphics[width=0.19\linewidth]{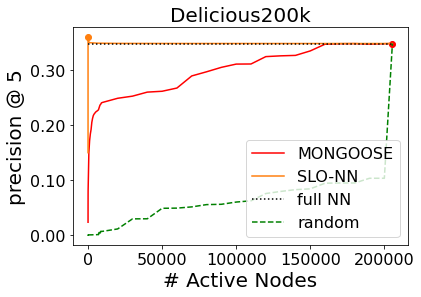}
\includegraphics[width=0.19\linewidth]{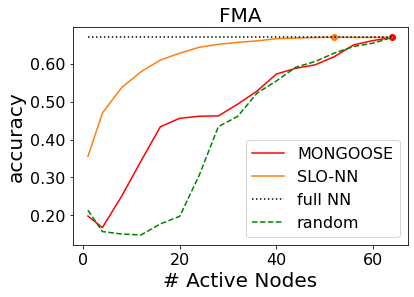}
\caption{Number of computed nodes per layer vs. accuracy. Yellow dots indicate where \slonns{} first reach maximum accuracy.}
\label{fig:approach_vs_random}
\end{figure*}

%\begin{figure*}[t]
%\centering 
%\includegraphics[width=0.19\linewidth]{figures/FMNIST/free_vs_cossim.png}
%\includegraphics[width=0.19\linewidth]{figures/Kitsune/free_vs_cossim.png}
%\includegraphics[width=0.19\linewidth]{figures/Wiki10/free_vs_cossim.png}
%\includegraphics[width=0.19\linewidth]{figures/AmazonCat13k/free_vs_cossim.png}
%\includegraphics[width=0.19\linewidth]{figures/Delicious200k/free_vs_cossim.png}
%\includegraphics[width=0.19\linewidth]{figures/FMAMusic/free_vs_cossim.png}

%\caption{Average number of dot products per layer versus accuracy. The number of computed dot products for a layer is equal to its corresponding number of computed hash functions plus its number of computed nodes.
%of hash functions plus the number of active nodes computed for the layer as each correspond to one dot product.
%}
%\label{fig:free_vs_cossim}
%\end{figure*}

\begin{figure*}
\centering 

\includegraphics[width=0.19\linewidth]{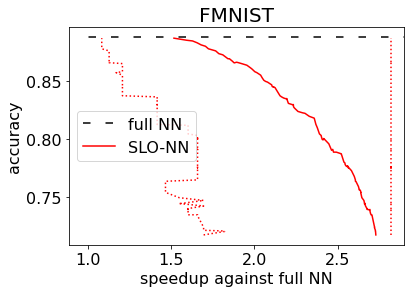}
\includegraphics[width=0.19\linewidth]{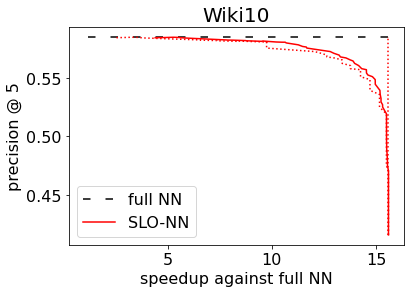}
\includegraphics[width=0.19\linewidth]{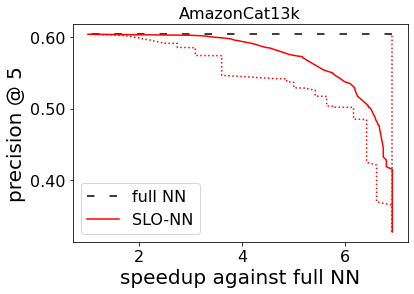}
\includegraphics[width=0.19\linewidth]{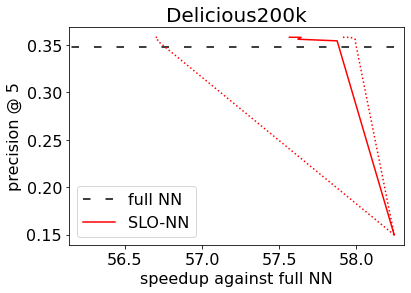}
\includegraphics[width=0.19\linewidth]{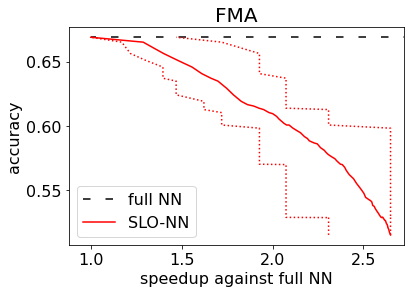}

\caption{Solid (resp. dotted) lines indicate the average (resp. min/max) speedup against the full neural network.
%(or mean speedup) while
% Dotted lines indicate the corresponding min/max number of active nodes.
%(or speedup).
Speedup depends on the number of computed nodes in the neural network for each data input. 
%Higher confidence allows the \slonn{} to drop out more nodes. 
%The technique is able to activate a subset of nodes and maintains high accuracy. Notice that a more lenient accuracy target allows for more speedup.
}
\label{fig:max_dropout}
\end{figure*}

\begin{figure*}
\centering 
\includegraphics[width=0.19\linewidth]{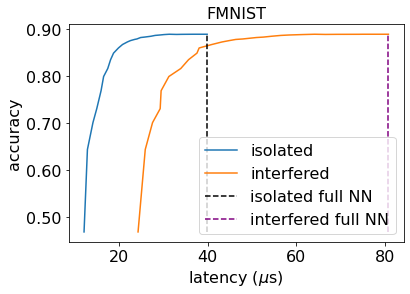}
\includegraphics[width=0.19\linewidth]{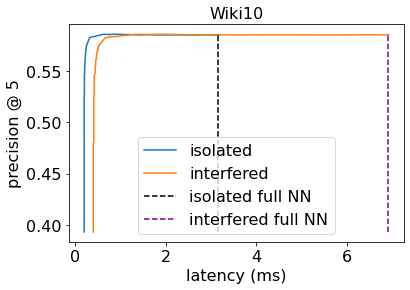}
\includegraphics[width=0.19\linewidth]{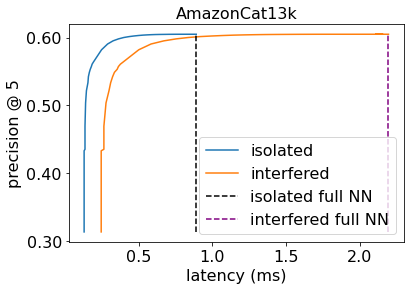}
\includegraphics[width=0.19\linewidth]{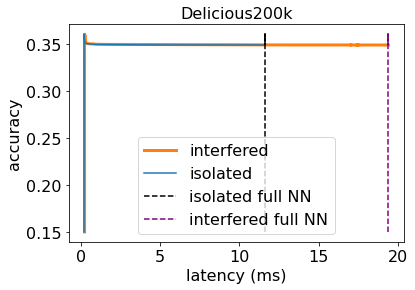}
\includegraphics[width=0.19\linewidth]{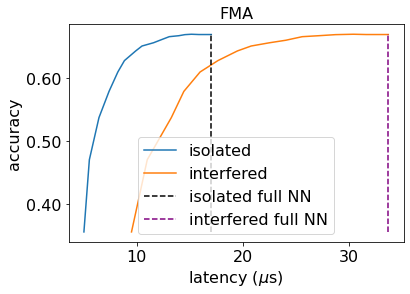}

\caption{Accuracy-latency tradeoff for LCAO \slonns{}. The \textit{isolated} curves show the performance of various \slonns{} when running alone on the worker machine. The \textit{interfered} curves show performance when a two instances of the same \slonn{} are co-located and queried simultaneously. 
%In general, \slonns{} can achieve \textit{uninterfered} inference latency during co-location.
%the machine while the curve labeled interfered exemplifies performance when its co-located with another running model instance. In most cases, the neural network can achieve an inference latency during isolation and during co-location, demonstrating it's ability to adapt to co-location interference while maintaining high accuracy.
}
\label{fig:min_dropout}
\end{figure*}

\noindent \textbf{Model Architectures and Datasets}
We evaluate \slonns{} on five datasets~\cite{xiao2017fashionmnist, fma_dataset, wiki10, amazon13k, delicious200k}, summarized in Table~\ref{tab:datasets}.
FMNIST is a multi-classification dataset of fashion products.
Kitsune is an anomaly detection dataset for detecting network attacks via packet statistics.
FMA is a music analysis dataset containing 106,574 tracks from 16,341 artists and 14,854 albums, arranged in a hierarchical taxonomy of 161 genres.
Wiki10 dataset is a collection of Wikipedia articles with associated user-defined tags formed from over 2 million Wikipedia articles.
AmazonCat-13K is a product-to-product recommendation dataset. 
Delicious-200K dataset is generated from a vast corpus of almost 150 million bookmarks from Social Bookmarking Systems.
Amazon-670K dataset is a product to product recommendation dataset.
%Wiki10, AmazonCat-13K, Delicious-200K, and Amazon-670K each exemplify a practical multi-label classification task.
% : FMNIST~\cite{xiao2017fashionmnist}, % FMA~\cite{fma_dataset}, 
% Kitsune~\cite{kitsune}, 
% Wiki10~\cite{wiki10}, AmazonCat-13K~\cite{amazon13k}, and Delicious-200K~\cite{delicious200k}.
% Table~\ref{tab:datasets} displays the train and test set sizes of the datasets as well as the neural network architecture for each dataset.

\noindent \textbf{Model Pruning} We statically neuron prune the baseline models architectures to ensure that each is reasonably sized for its corresponding dataset~\cite{neuronpruning,blalock2020state}.
%We argue that each of the models are reasonably sized for each dataset.
For the FMNIST and FMA models, we applied neuron model pruning ~\cite{neuronpruning,blalock2020state} prior to transforming them into \slonns{}.
%to minimize the number of neurons in the neural network prior to integrating them into the \slonn{}. 
\slonns{} for Wiki10, Delicious200k, and Amazoncat13k, feature a Node Activator at the output layer only; these models are not pruned since pruning cannot impact the output layer.
%the output layer and the neurons of the output layer cannot be statically pruned.
%Further, each of the neural networks have similar number of neurons compared to the models used in real inference serving systems~\cite{deeprecsys}. 

\noindent \textbf{Inference Platform}
Our evaluation is conducted on a server equipped with two 64-core Intel Xeon Gold 6226R CPUs.
Most inference serving systems employ server/edge CPUs due to their abundance and cost-efficiency in comparison to GPUs~\cite{park2018deep,hazelwood:2018:mlatfb}. We plan to evaluate on GPUs in future work.
%We show results employing both single core and multi-core to demonstrate the method's effectiveness in both sequential and parallel execution environments. 
%In practice, the number of cores assigned to an inference model depends on the size of the model and the SLOs. 
%For the smaller neural networks in our experiments (used for the FMNIST, FMA, Wiki10, and Amazoncat-13K datasets), we assign one core. 
%For the FMNIST, FMA, Wiki10, and AmazonCat13k models, we assign one core.
%For the Delicious-200K model, we assign four cores.

\noindent \textbf{\slonn{} Implementation}
Our implementation of \slonns{} use NumPy 1.19.5~\cite{numpy}. Numba 0.53.1~\cite{numba} is used to compile into fast machine code.
%implemented the SLO-Aware NN framework with Numpy and used Numba to compile into fast machine code. 
%Fig.~\ref{fig:latency_breakdown} compares our implementation with PyTorch~\cite{pytorch}.
%to demonstrate effectiveness against existing inference execution implementations.
Fig.~\ref{fig:latency_breakdown} compares the run times of the activating the entire neural network with PyTorch and the \slonn{}.
All bars represent full forward pass (i.e., all nodes computed) median latencies over 100 runs of the evaluated neural networks. \textit{PyTorch} bars represent the inference latency of out-of-the-box PyTorch. 
 Fig.~\ref{fig:latency_breakdown}  demonstrates that \slonns{} exhibit low overhead even if no computation is dropped out.
%\slonn{} and \slonn{}+Agg illustrate the extra overhead of LSH-related computations with and without aggregation (\S\ref{sec:node-confidence}), respectively.
%Note that \slonn{} performs similarly to PyTorch. 
%Moreover, omitting the aggregation step significantly reduces the performance overhead of LSH.
%In this paper, we focus on online inference, and plan to investigate batch inference in future work.
In this paper, we focus on latency-critical online inference where batch inference is often too slow (e.g., due to queuing delays \cite{park2018deep}). Many real-time inference systems implement a batch size of 1 \cite{park2018deep,choy_2020,machynist_kippinitreal_2020} and most are restricted to a small batch size \cite{park2018deep,hazelwood:2018:mlatfb}. We plan to investigate batch inference in future work.

\section{Preliminary Results}
\label{sec:results}

\subsection{\slonns{} vs. existing dropout frameworks}
\label{sec:accuracy-results}
Fig.~\ref{fig:approach_vs_random} showcases the ability of \slonns{} to select the \textit{most important} nodes---those that optimize accuracy---to serve inference queries when performing dropout. The x-axes represent the number of nodes computed during an inference query. The y-axes report inference accuracy, averaged across all test set samples. Fig.~\ref{fig:approach_vs_random} compares three dropout schemes---\slonn{}, \textsc{Mongoose}, and \textit{random}---to the baseline accuracy of the full neural network (where all nodes are computed). %\slonn{} is our proposed implementation which features the \freehash{} LSH hash family. 
%The selected LSH hyper-parameters are $K=10$ and $L=2$. Unless otherwise stated, this \slonn{} implementation is constant throughout our experiments. 
\textsc{Mongoose} is the most similar prior work to \slonns{} which proposes LSH-based dropout at training ~\cite{MONGOOSE}.  
%The \textit{random} scheme drops out nodes uniformly at random for each input.

Fig.~\ref{fig:approach_vs_random} shows that for each dataset, \slonns{} significantly outperform \textsc{Mongoose} and \textit{random} dropout.
Given the same number of active nodes, the \slonn{} is up to 50\% more accurate than \textsc{Mongoose}.
We expect this discrepancy is due to the differing LSH training procedures of \slonns{} and \textsc{Mongoose}. 
Specifically, \textsc{Mongoose} never realizes the entire activation of a data input in order to achieve faster forward propagation and gradient update, given their goal of \textit{training}.
\textsc{Mongoose} only considers \textit{subsets} of node activations when training its LSH, which leads to imprecise node importance ranks.
Training can tolerate and adapt to inaccurate node importance ranks since the inaccuracy emulates random adaptive dropout, whereas inference requires node importance lists to have higher degrees of precision.
\slonns{} leverage complete node activations during LSH training to establish node importance, which results in better accuracy.

\slonns{} quickly reach and sustain full neural network accuracy with as few as 0.01\% of the total nodes and as many as 94\%. The point at which maximum accuracy is achieved is marked with a yellow dot in each graph. 
Interestingly, for Delicious200k, \slonns{} achieve higher accuracy than the original neural network when computing only 0.01\% of the nodes. Its accuracy converges to that of the full neural network with more computed nodes.
% significantly less than the total number of nodes in the network.
Overall, \slonns{} effectively identify and selectively compute the most important nodes in a neural network on a \textit{per-input basis.}

\subsection{\slonns{} with an ACLO Optimization Target}
The ACLO optimization target directs an \slonn{} to minimize inference latency given an accuracy SLO target. For a given input example ACLO involves minimizing the number of computed nodes given an accuracy constraint. In theory, ``easy'' inputs can be computed faster than ``hard'' ones.
Fig.~\ref{fig:max_dropout} compares inference speedup of \slonns{} over a full neural network (x-axes) with the achieved accuracy (y-axis). Specifically, these experiments are the result of supplying \slonns{} with an accuracy target and asking it to minimize inference latency (i.e., minimize the number of computed nodes) for each input example in the test set. 
%the accuracy versus the speedup against the full neural network in Fig.~\ref{fig:max_dropout}.
% We implement \slonns{} as in \S\ref{sec:accuracy-results}.
% with \freehash{} and hyper-parameter settings $L=2$ and $K=10$.

Fig.~\ref{fig:max_dropout} shows the minimum (left curve), average (middle curve), and maximum (right curve) achieved by \slonns{} at various accuracy targets. Overall, \slonns{} exhibit a high range of inference speedup. For example, for a high accuracy target (within 0.3\% accuracy of the full neural network), \slonns{} exhibit speedups of $1.1-2.8\times$ for FMNIST, $8.4-15.6\times$ for Wiki10, $1.8-6.9\times$ for AmazonCat13k, $56.7-57.9\times$ for Delicious200k, and $1.2-1.7\times$ for FMA.
% Even for the highest accuracy targets, 
% Observe that the number of active nodes varies and exhibits a wide range even for high accuracy. 
%This result further exemplifies the idea that some inputs are easier to perform inference on than others.
%Moreover, \slonns{} exhibit average speedups of $1.6\times$ for FMNIST, $9.6\times$ for Wiki10, $3.3\times$ for AmazonCat13k, $56.7\times$ for Delicious200k, and $1.3\times$ for FMA.

%Further, Fig.~\ref{fig:max_dropout} illustrates that \slonns{} can also simultaneously serve a range of accuracy targets.
%For instance, given an accuracy target 5\% below the full neural network accuracy, \slonns{} achieve even greater average speedups of $2.1\times$ for FMNIST, $11.7\times$ for Wiki10, $5.9\times$ for AmazonCat13k, $58.1\times$ for Delicious200k, and $1.8\times$ for FMA.
Overall, \slonns{} are able to achieve significant latency improvements while retaining accuracy.
% by dynamically adapting dropout on a per-query basis.
% A wide range indicates that the model is able to dropout a significant amount of nodes for examples where the network is confident.
% Further, the approach is able to activate significantly less nodes than the total number of nodes in the original neural network while achieving similar or better accuracy.

\subsection{\slonns{} with an LCAO Optimization Target}
The LCAO optimization target directs an \slonn{} to minimize dropout (so as to optimize accuracy) given a latency target.
Furthermore, LCAO takes into consideration information about current machine utilization and pre-computed Latency Profiles (Fig.~\ref{fig:overall_LSH}) to account for the effects of intermittent co-location interference on latency.
%to framework takes as input an latency constraint and minimizes dropout needed to satisfy the latency SLO.
%The framework leverages a latency profile of the neural network to anticipate the effects of intermittent co-location interference on latency.
Fig.~\ref{fig:min_dropout} compares inference latency of \slonns{} (x-axes) with inference accuracy (y-axes) when operating under the LCAO optimization target. The dotted black and purple vertical lines show full model neural network inference latency when inference is run in isolation versus when it experiences co-location interference, respectively. Here, our co-location interference scenario considers a second co-located copy of the same inference model, serving back-to-back inference requests.
The blue/orange curves illustrate the accuracy latency tradeoff for \slonns{} in isolated/interfered execution scenarios.
% , respectively. 

%During co-location, the full neural network latency (comparing black to purple lines) increases by $1\times$ for FMNIST, $1.2\times$ for Wiki10, $1.5\times$ for AmazonCat13k, $0.7\times$ for Delicious200k, and $1\times$ for FMA.
Notably, Fig.~\ref{fig:min_dropout} demonstrates that zero latency degradation (with respect to full network latency) can be achieved by \slonns{} \textit{when interfered}.
%In particular, the point at which the orange curves intersect with the dotted black vertical lines indicate the accuracy that can be achieved for interfered inference while retaining the latency of un-interfered inference. 
%\slonns{} for Wiki10 and Delicious200k achieve zero loss in accuracy while retaining the latency of un-interfered inference.
Wiki10, Delicious200k, FMNIST, AmazonCat13k, and FMA exhibit $0\%$, $0\%$, $2.1\%$, $0.3\%$, and $4.1\%$ accuracy drop compared to the full neural network, respectively, while retaining the latency of un-interfered inference.
%Further, for more lenient latency SLOs the \slonns{} can achieve higher accuracy.
%For instance, given that the latency SLO can tolerate a latency increase of $0.2\times$ above the latency in isolation, under co-location interference, FMNIST, AmazonCat13k, and FMA exhibit $1\%$, $0.1\%$, and $1.8\%$ accuracy drop compared to the full neural network, respectively.
%Moreover, under queueing delays from high query load, the \slonn{} may drop computation to meet the latency deadline in both isolated and interfered scenarios.
Overall, Fig.~\ref{fig:min_dropout} illustrates that the LCAO \slonn{} is able to simultaneously meet a range of latency SLOs even under intermittent co-location interference while maintaining high accuracy.

\section{Related Work}
\label{sec:related}

%\noindent \textbf{Optimizing resource utilization}
%Inference serving systems often aim to keep resource utilization high for cost-effectiveness.
%Techniques to maximize utilization include INFaaS~\cite{romero:infaas} which automatically selects models and worker machines to ensure that SLOs are met while keeping machine utilization high. \cite{horus} introduces an interference-aware scheduler for training and serving inference on GPUs which attempts to place models onto GPUs such that resource utilization is maximized.

%\noindent \textbf{Optimizing inference latency}
%\cite{willump} proposes an automatic ML model optimizer which given an accuracy target, trains an approximate model with a subset of the input features and then identifies and classifies "easy" data inputs with the low latency approximate model while leaving "hard" inputs to a slower but more accurate model. Willump targets inference workloads that are bottlenecked by feature extraction. In this work, we focus on inference. 

%Model pruning is a popular technique employed to reduce latency by permanently removing weights of the neural network~\cite{blalock2020state}. However, model pruning often comes at the cost of reducing the accuracy of the model.
\noindent \textbf{Inference Serving Systems}
Most modern model serving systems (e.g., Clipper~\cite{crankshaw2017clipper}, Amazon Sagemaker, Microsoft AzureML, INFaaS~\cite{romero:infaas}, Horus~\cite{horus}, Perseus~\cite{lemay2020perseus}) treat ML inference as a black box.
These approaches must train and manage many models to meet diverse SLOs under varying query loads.
%Further, these approaches rely on loading a new model to match the SLOs for each query.
As observed by \cite{lemay2020perseus}, model load times are often significantly longer (up to 100$\times$ slower) than inference; thus, switching models online is likely to lead to latency SLO violations.
%These techniques are especially vulnerable to SLO violations due to volatile query loads where they either under-utilize the worker machines at low load or violate SLOs from bursty traffic.
%Nor do they anticipate the effects of co-location on inference latency.
\slonns{} circumvent these issues by managing a single model which can dynamically adapt to a diversity of SLOs, changing query load, and co-location interference.

\noindent \textbf{ML Inference Optimizations}
Model pruning is a popular technique employed to compress a neural network by permanently removing connections between neurons or the neurons themselves and often incurs accuracy loss~\cite{blalock2020state,neuronpruning,weightpruning}. 
Static model pruning is oblivious to the notion that some input queries are ``easy'' and thus cannot leverage per query activation sparsity for inference acceleration.
\slonns{} are complementary to static model pruning as the framework can take as input a statically pruned model. 
%Quantization has been proposed to compress a neural network where weights of the neural network are represented as integers instead of floats~\cite{jacob2017quantization}. This optimization is complementary to \slonns{}.

\noindent \textbf{LSH for Neural Networks}
\textsc{Reformer}~\cite{reformer} propose an LSH-based transformer model where they replace the attention layers of a transformer model with LSH tables to produce a more compressed transformer model. Their technique is only applicable to transformer models, and requires the LSH-based transformer model to be trained with the LSH tables.
\textsc{Mongoose}~\cite{MONGOOSE} applies an LSH-based dropout-at-training scheme to speed up neural network training. The technique is an extension of prior work which maps adaptive dropout as a maximum inner product search problem~\cite{slide,spring2016scalable}. \textsc{Mongoose} only considers partial node activation when training its LSH which leads to inaccurate node importance ranks. \slonns{} take into account the full node activation at LSH training which, as we demonstrate, leads to significant improved performance.

\noindent \textbf{Dynamic Neural Networks}
Recent work has proposed
%Many prior work has been proposed in designing
dynamic neural network architectures which exhibit conditional computation based on SLO targets~\cite{hua2019channel,gao2019dynamic,yu2018slimmable}. However, these designs restrict either the model architecture or training procedure, and may not achieve state-of-the-art accuracy ~\cite{hua2019channel,gao2019dynamic,yu2018slimmable}. In contrast, \slonns{} make no such restrictions as it provides a general framework to develop facilities for conditional computations \textit{at inference}.

\section{Future Work and Conclusions}
%\noindent \textbf{Extension to other types of neural network layers}
In this paper, we focus on latency-critical online inference for Multi-Layer Perceptrons networks. In ongoing work, we are investigating the application \slonns{} to other architectures, such as convolutional neural networks.
We also plan to investigate batch inference for \slonns{}.
Many solutions to batch inference with \slonns{} are possible, such as using LSH to cluster batch inputs into parallel micro-batches or dividing the selected nodes across inputs according to a weighting scheme that accounts for input difficulty. Scheduling with \slonn{} batch inference is difficult as it requires making adaptive batch size decisions under varying co-location interference, queuing delay, and query load.

% % \noindent \textbf{Batch inference}
% Here we focus on latency-critical online inference and we intend to adapt our framework to batch inference as well.

% \noindent \textbf{Larger scale model co-location deployment}
Our current experiments evaluate \slonns{} running on CPUs with a maximum of two co-located models. We plan to scale up our evaluation by adding more complex query patterns, co-location configurations, and hardware platforms.
Along these lines, we are also interested in understanding how our \slonns{} can be to designed to accelerate inference under shifting query data distributions by employing lightweight online updates to the Node Activator.

% \noindent \textbf{Other Node Activator designs}
Finally, while the Node Activator in \slonns{} is based on
% upon
LSH, and we plan to study other node ranking mechanisms.

% \noindent \textbf{Node Activator adaptivity to data distribution shift}

% \section{Conclusions}
% \label{sec:conclusions}
In summary, we present \slonns{} as a type of neural network which can dynamically adapt inference computation according to SLO optimization targets and co-location interference on a \textit{per-query basis}. \slonns{} place no restrictions on training, enable a variety of SLOs to be met with just a single model, and exhibit benefits beyond what can be achieved by static model pruning techniques.
%We show that \slonns{} achieve average speedups ranging $1.3-56.7\times$ with zero or negligible accuracy difference (less than 0.3\%) compared to the original neural network.
%We also exemplify \slonns{} are capable of simultaneously serving a range of accuracy targets while minimizing latency, where lowering an accuracy target results in significant increased speedups.
%We demonstrate that \slonns{} are able to proactively alleviate latency degradation from co-location interference while maintaining high accuracy.

\bibliographystyle{ACM-Reference-Format}
\bibliography{new_main}

\end{document}